\documentclass[conference]{IEEEtran}
\IEEEoverridecommandlockouts
\usepackage{cite}
\usepackage{amsmath,amssymb,amsfonts}
\usepackage{algorithmic}
\usepackage{graphicx}
\usepackage{textcomp}
\usepackage{xcolor}
\usepackage{url}
\usepackage{multirow}
\usepackage[whole]{bxcjkjatype}
\def\BibTeX{{\rm B\kern-.05em{\sc i\kern-.025em b}\kern-.08em
    T\kern-.1667em\lower.7ex\hbox{E}\kern-.125emX}}
\begin{document}

\title{Summarization of Investment Reports Using Pre-trained Model
}

\author{\IEEEauthorblockN{Hiroki Sakaji}
\IEEEauthorblockA{\textit{School of Engineering} \\
\textit{The University of Tokyo}\\
Tokyo, Japan \\
sakaji@sys.t.u-tokyo.ac.jp}
\and
\IEEEauthorblockN{Ryotaro Kobayashi}
\IEEEauthorblockA{\textit{School of Engineering} \\
\textit{The University of Tokyo}\\
Tokyo, Japan \\
}
\and
\IEEEauthorblockN{Kiyoshi Izumi}
\IEEEauthorblockA{\textit{School of Engineering}\\
\textit{The University of Tokyo}\\
Tokyo, Japan \\
}
\and
\IEEEauthorblockN{Hiroyuki Mitsugi}
\IEEEauthorblockA{\textit{Data-driven Science Dept.} \\
\textit{Frontier Technologies Research Center.} \\
\textit{Daiwa Institute of Research Ltd.}\\
Tokyo, Japan \\
}
\and
\IEEEauthorblockN{Wataru Kuramoto}
\IEEEauthorblockA{\textit{Fund Management Division} \\
\textit{Daiwa Asset Management Co.Ltd.}\\
Tokyo, Japan \\}
}

\maketitle

\begin{abstract}
\renewcommand{\thefootnote}{\fnsymbol{footnote}}
\footnote[0]{© 2023 IEEE.  Personal use of this material is permitted.  Permission from IEEE must be obtained for all other uses, in any current or future media, including reprinting/republishing this material for advertising or promotional purposes, creating new collective works, for resale or redistribution to servers or lists, or reuse of any copyrighted component of this work in other works.}
\renewcommand{\thefootnote}{\arabic{footnote}}
In this paper, we attempt to summarize monthly reports as investment reports.
Fund managers have a wide range of tasks, one of which is the preparation of investment reports.
In addition to preparing monthly reports on fund management, fund managers prepare management reports that summarize these monthly reports every six months or once a year.
The preparation of fund reports is a labor-intensive and time-consuming task.
Therefore, in this paper, we tackle investment summarization from monthly reports using transformer-based models.
There are two main types of summarization methods: extractive summarization and abstractive summarization, and this study constructs both methods and examines which is more useful in summarizing investment reports.
\end{abstract}

\begin{IEEEkeywords}
Summarization, Pre-trained Model, Investment Report
\end{IEEEkeywords}

\section{Introduction}
\begin{figure}[h]
 \begin{center}
 \includegraphics[width=\hsize]{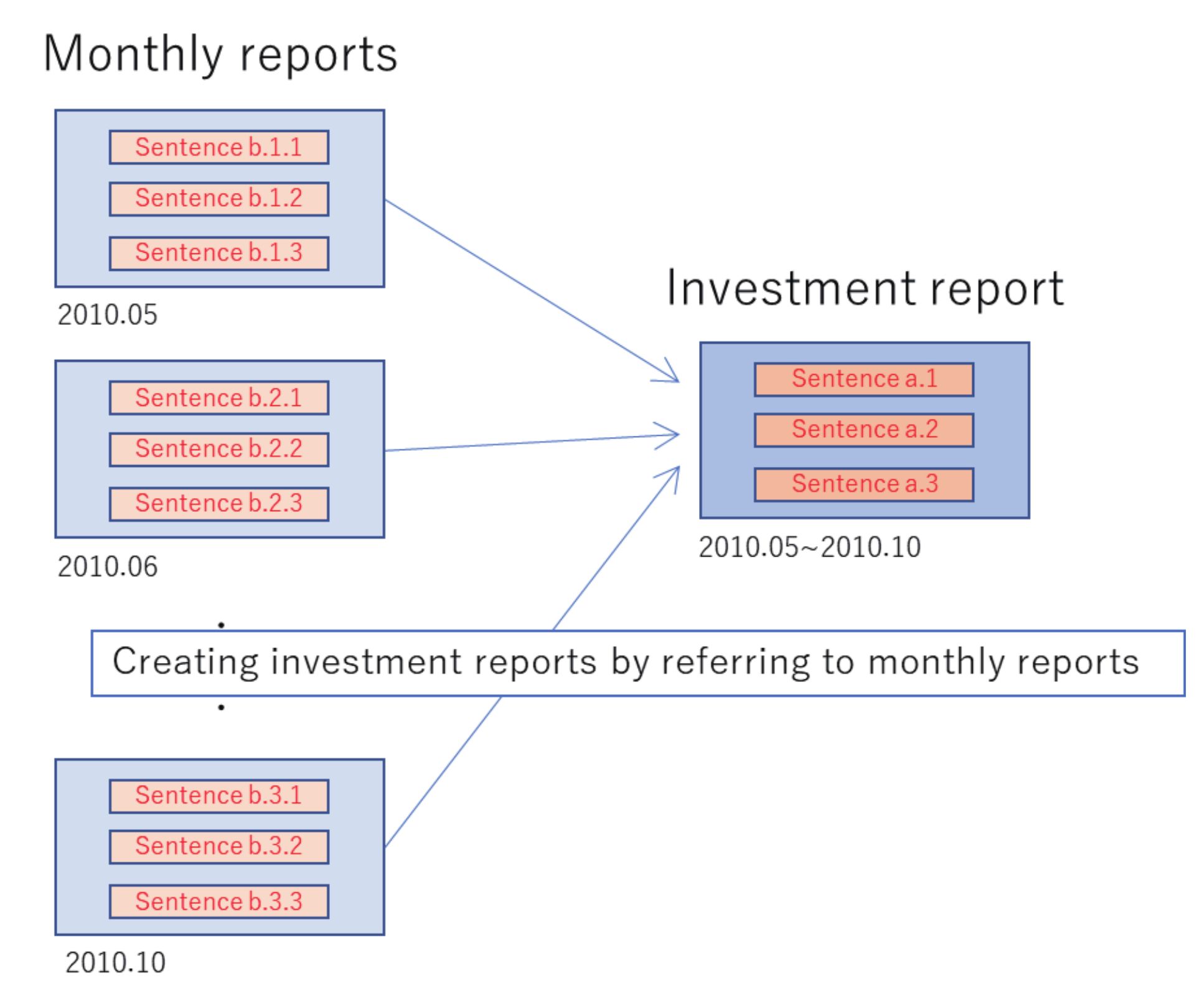}
 \end{center}
 \caption{Investment reports.}
 \label{img:investment_reports}
\end{figure}

There are a variety of finance-related occupations in the world, including fund managers, who manage funds.
Fund managers have a wide range of tasks, one of which is the preparation of investment reports.
In addition to preparing monthly reports on fund management, fund managers prepare management reports that summarize these monthly reports every six months or once a year.
The preparation of fund reports is a labor-intensive and time-consuming task.
The investment report is prepared with reference to the monthly report as shown in Figure \ref{img:investment_reports}.
However, the investment reports do not completely match the monthly reports, as only sentences containing important information contained in each monthly report are used, or multiple sentences are combined and generated as a new sentence.
It is also possible that information not included in the monthly report will be added to the investment report.

Meanwhile, AI in language processing is growing along with the evolution of neural networks.
In particular, transformer-based\cite{vaswani2017attention} models such as BERT\cite{Devlin2018} have performed very well and have been used for various applications.
In recent years, interactive AIs such as ChatGPT have emerged and spread into the real world, being incorporated into various applications.
Therefore, in this paper, we tackle investment summarization from monthly reports using transformer-based models.
Specifically, we challenge a multi-document summary, taking the monthly report as input and outputting the investment report.
There are two main types of summarization methods: extractive summarization and abstractive summarization, and this study constructs both methods and examines which is more useful in summarizing investment reports.

The contributions of this study are as follows.
\begin{itemize}
    \item We compared extractive and abstractive summarization in a summarization task in investment reports and showed that abstractive summarization is more effective.
    \item By analyzing the summarization performance for each type of fund, we identified the types that are easy or difficult to summarize and discussed the reasons for this.
\end{itemize}

\section{Overview of our summarization}
\begin{figure}[h]
 \begin{center}
 \includegraphics[width=\hsize]{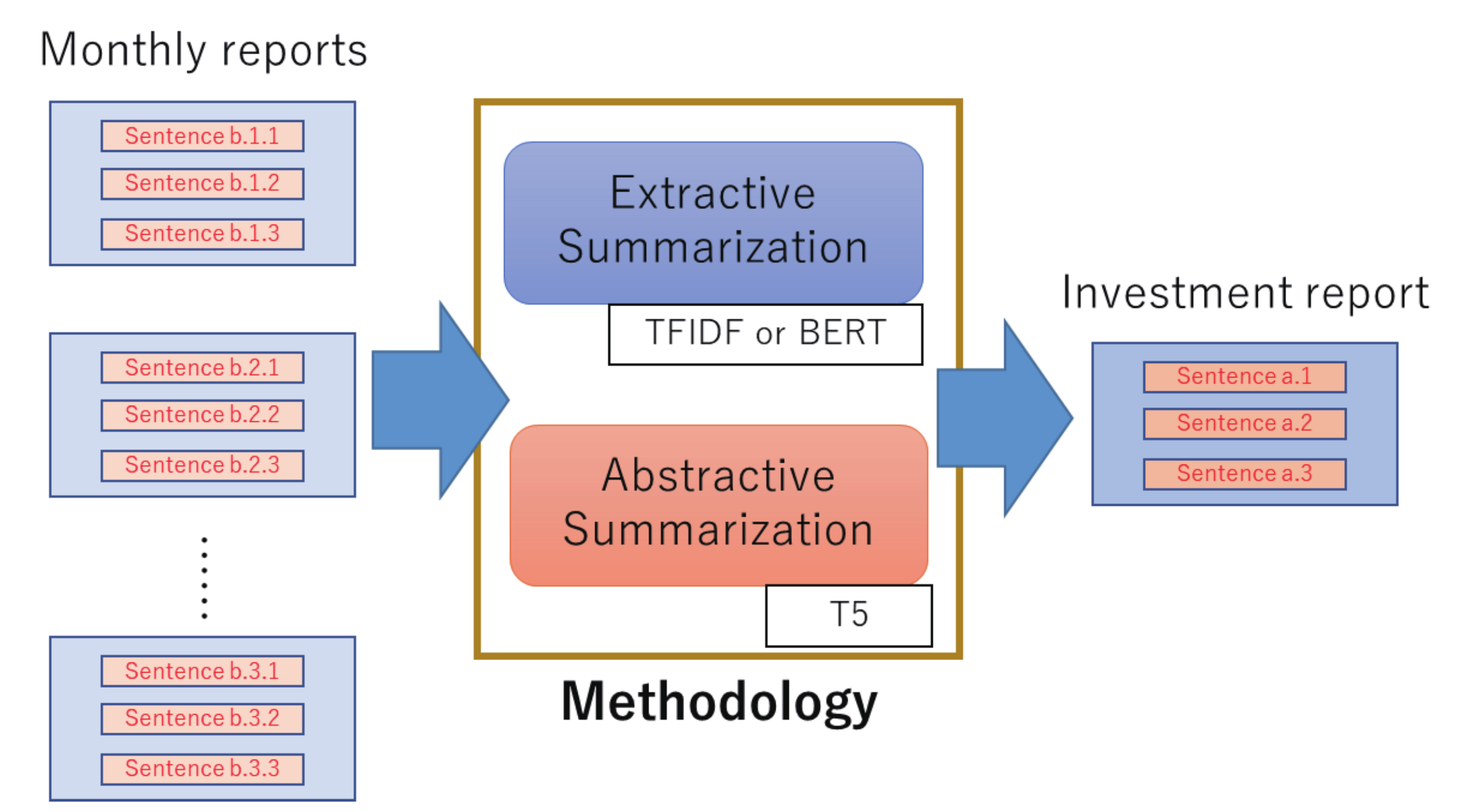}
 \end{center}
 \caption{The flow of report summarization.}
 \label{img:flow}
\end{figure}

In this section, we introduce how to summarize investment reports from monthly reports.
From Figure \ref{img:investment_reports}, investment reports refer to monthly reports.
Therefore, we use monthly reports as input data for generating investment reports.
Figure \ref{img:flow} shows the flow of investment reports summarization.

From Figure \ref{img:flow}, we try two different summarization methodologies, extractive summarization, and abstractive summarization, in this paper.
For creating extractive summarization training data, we use two types of methods.
One is the TFIDF base method, and the other is the BERT base method.
These methods are introduced in section \ref{sq:extractive}.
On the other hand, we use T5\cite{2020t5} for abstractive summarization.
Table \ref{tb:unyo} shows an example of an investment report.
Table \ref{tb:monthly} shows an example of a monthly report associated with the investment report in Table \ref{tb:unyo}.
These examples deal with topics related to Russian market conditions.
\begin{table}[h]
\centering
\caption {Example of investment report.}
\small
\begin{tabular}{p{8.4cm}}
\hline
ロシア株式市況は、現地通貨建てでほぼ横ばいとなりました。欧米との外交問題、特に2018年4月前半に米国からの追加経済制裁が発表 されたことから、一時大きく下落しましたしかし、原油価格が高水準で推移したことはプラス材料となりました。\\
(The Russian stock market was mostly flat in local currency terms. Diplomatic issues with Europe and the U.S., especially the announcement of additional economic sanctions from the U.S. in the first half of April 2018, led to a large temporary decline; however, the high level of oil prices was a positive factor.)
\\ \hline
\end{tabular}
\label{tb:unyo}
\end{table}
\begin{table}[h]
\centering
\caption {Example of mothly report.}
\small
\begin{tabular}{p{8.4cm}}
\hline
ロシア株式市場は上昇（現地通貨ベース）2018年2月のロシア株式市場は上昇しました。月前半は、米国の長期金利が急 上昇したことや原油価格の急落、並びに市場のボラティリティ上昇に伴って投資家のリスク許容度が低下しリスク資産に対する売り圧力が高まったことなどが嫌気され、市場は下げ幅を拡大しました。しかしながら月後半には、金融市場が落ち着きを取り戻すにつれ、原油価格が戻り歩調となったことなどが好感され、徐々に値を戻す展開となりました。また、大手格付け会社がロシアの信用格付けを2006年以降で初めて引き上げ、投資適格級としたと報じられたことも、相場の支援材料となりました。 \\
(Russian Stock Market Rises (Local Currency Basis) The Russian stock market rose in February 2018. In the first half of the month, the market was down due to a sharp rise in long-term U.S. interest rates, a sharp drop in oil prices, and increased selling pressure on risk assets as investors' risk tolerance declined due to increased volatility in the market. In the second half of the month, however, as financial markets regained their composure, the price of crude oil returned to its previous highs, and the market gradually recovered. The market was also supported by reports that a major credit rating agency raised Russia's credit rating to investment grade for the first time since 2006.)
\\ \hline
\end{tabular}
\label{tb:monthly}
\end{table}

\section{Extractive Summarization}
\label{sq:extractive}
Our extractive summarization method involves automatically extracting sentences from multiple monthly reports with the aim of generating a summary that can substitute the corresponding management reports for the given period.
We partition the management reports and monthly reports into training and test data sets, and construct a model of extractive summarization using machine learning-based methods as follows.
First, we label each sentence of the monthly report in the training data set to determine whether it is plausible to include it in the summary. 
For this, we use a binary labeling approach based on whether or not the sentence has a high degree of similarity to one of the corresponding management reports for the period. 
We measure similarity by the cosine similarity of the vector representation of the sentences. 
We create the vector representation of each sentence in the management reports and monthly reports in two ways: by vectorizing the sentences using TFIDF and by creating an embedded representation using BERT.
We use a base-size BERT model that was pre-trained on Japanese texts, which were published by the Inui Lab at Tohoku University\footnote{\url{https://huggingface.co/cl-tohoku/bert-base-japanese-whole-word-masking}}.
Next, we build the model by fine-tuning the BERT model as a binary classification task using the created training data. 
Predictions are made by applying a linear layer to the output of the final layer of the BERT model.
We use the same BERT model as in the previous step and only fine-tune the upper layer of the Transformer layer.
We obtain the number of epochs as 9 and the batch size as 4 through Bayesian optimization using Optuna\footnote{\url{https://optuna.org/}}.
Finally, we create a summary by connecting multiple sentences from the top that are estimated to have the highest likelihood of being included in the summary.

\begin{figure}[h]
 \begin{center}
 \includegraphics[width=\hsize]{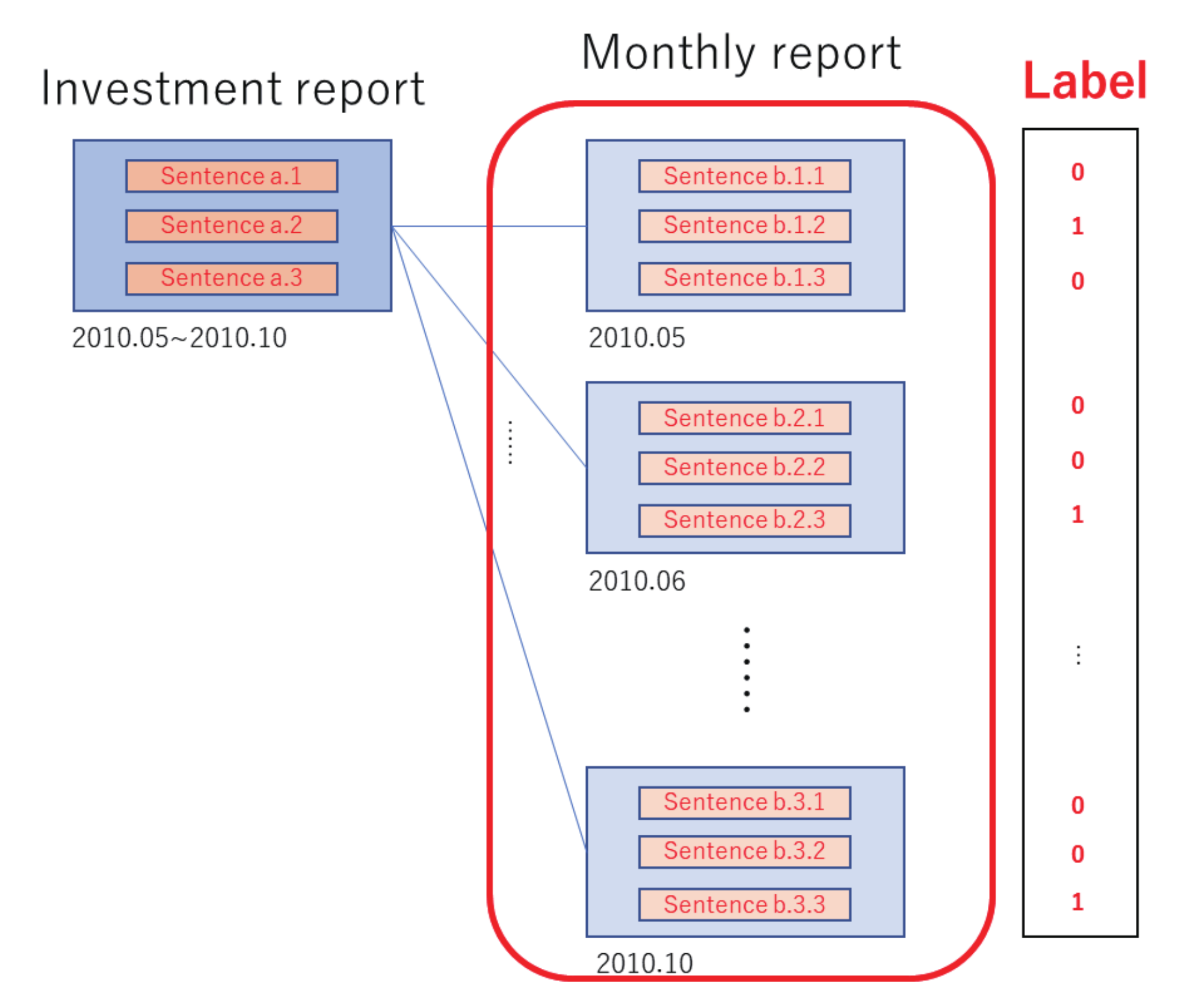}
 \end{center}
 \caption{Labeling for extractive summarization.}
 \label{img:labeling}
\end{figure}

\section{Abstractive Summarization}
In this section, we introduce abstractive summarization using T5.
We use t5-base-japanese\footnote{\url{https://huggingface.co/sonoisa/t5-base-japanese}} as T5 model. 
We decide the input size is 1024.
This value is slightly higher than the average of the 6 or 12 monthly reports.
Additionally, we decide the output size is 256.
This value is slightly higher than the average of the investment reports.
In this paper, the epoch number is 10, and the batch size is 8.
These values are obtained by grid search.
Specifically, we use T5 of transformers\footnote{\url{https://huggingface.co/docs/transformers/model_doc/t5}} for the implementation of our abstractive summarization model. 

\section{Experiment}
We evaluate our extractive summarization methods and abstractive summarization method using investment reports and monthly reports.
We utilize 2,631 Japanese investment reports and 18,953 Japanese monthly reports tied to these investment reports.
70\% of these reports are used for training data, 10\% for validation data, and 20\% for test data.
We evaluate below three methods.
Ex-BERT is BERT based extractive summarization method.
Ex-TFIDF is TFIDF based extractive summarization method.
Ab-T5 is T5 based abstractive summarization method.
Specifically, we evaluate these methods using ROUGE-1, ROUGE-2, and ROUGE-L.
These methods are developed using PyTorch\footnote{\url{https://pytorch.org/}} and transformers\cite{wolf-etal-2020-transformers}.
Additionally, we also evaluate each type of fund.
Fund types include stock, bond, real estate, asset combination, and others.
Furthermore, there is also a classification of which region the fund targets: domestic, foreign, or domestic and foreign.
By evaluating each of these detailed categories, we find the types of investment reports that are suited for summarization.

\subsection{Experiment Results}
Table \ref{tb:results} shows the results of the experiment.
Additionally, Table \ref{tb:Ex-BERT}, Table \ref{tb:Ex-TFIDF}, and Table \ref{tb:Ab-T5} show the evaluation results for each type.
In Table \ref{tb:Ex-BERT}, Table \ref{tb:Ex-TFIDF}, and Table \ref{tb:Ab-T5}, D indicates domestic, F indicates foreign, and DF indicates domestic and foreign.
Moreover, we show examples of summarization results in Table \ref{tb:outputs}.
In Table \ref{tb:outputs}, outputs of Ab-T5 and Ex-TFIDF and the correct answer of summarization are indicated. 
\begin{table}[ht]
  \caption{Experiment results.}
  \begin{center}
  \small
  \begin{tabular}{l|r|r|r}
  \hline
  & ROUGE-1 & ROUGE-2 & ROUGE-L \\ \hline \hline
  Ex-BERT & 0.483 & 0.224 & 0.265 \\ \hline
  Ex-TFIDF & 0.497 & 0.233 & 0.274 \\ \hline
  Ab-T5 & {\bf 0.704} & {\bf 0.548} & {\bf 0.595} \\ \hline
  \end{tabular}
  \end{center}
  \label{tb:results}
\end{table}

\begin{table}[ht]
\centering
\caption{Detailed results in Ex-BERT. D indicates domestic, F indicates foreign, and DF indicates domestic and foreign.}
\begin{tabular}{|c|c|r|r|r|} \hline
	\multicolumn{2}{|c}{} & \multicolumn{1}{|c}{ROUGE-1} & \multicolumn{1}{|c}{ROUGE-2} & \multicolumn{1}{|c|}{ROUGE-L} \\ \hline
	\multicolumn{2}{|c|}{All} & 0.483 & 0.224 & 0.265 \\ \hline
	\multirow{3}{*}{Stock} & D & 0.443 & 0.191 & 0.242 \\ \cline{2-5}
	 & F & 0.506 & 0.233 & 0.267 \\ \cline{2-5}
	 & DF & 0.427 & 0.193 & 0.212 \\ \hline
	\multirow{2}{*}{Bond} & F & 0.476 & 0.243 & 0.246 \\ \cline{2-5}
	 & DF & 0.404 & 0.136 & 0.241 \\ \hline
	\multirow{3}{*}{Other} & D & 0.470 & 0.208 & 0.254 \\ \cline{2-5}
	 & F & 0.491 & 0.238 & 0.276 \\ \cline{2-5}
	 & DF & 0.498 & 0.230 & 0.269 \\ \hline
	\multirow{2}{*}{Asset combination} & D & 0.308 & 0.145 & 0.210 \\ \cline{2-5}
	 & DF & 0.532 & 0.172 & 0.242 \\ \hline
	Real estate & D & 0.514 & 0.254 & 0.250 \\ \hline
\end{tabular}
\label{tb:Ex-BERT}
\end{table}

\begin{table}[ht]
\centering
\caption{Detailed results in Ex-TFIDF. D indicates domestic, F indicates foreign, and DF indicates domestic and foreign.}
\begin{tabular}{|c|c|r|r|r|} \hline
	\multicolumn{2}{|c}{} & \multicolumn{1}{|c}{ROUGE-1} & \multicolumn{1}{|c}{ROUGE-2} & \multicolumn{1}{|c|}{ROUGE-L} \\ \hline
	\multicolumn{2}{|c|}{All} & 0.497 & 0.233 & 0.274 \\ \hline
	\multirow{3}{*}{Stock} & D & 0.465 & 0.197 & 0.252 \\ \cline{2-5}
	 & F & 0.503 & 0.225 & 0.257 \\ \cline{2-5}
	 & DF & 0.421 & 0.160 & 0.224 \\ \hline
	\multirow{2}{*}{Bond} & F & 0.616 & 0.305 & 0.328 \\ \cline{2-5}
	 & DF & 0.445 & 0.123 & 0.221 \\ \hline
	\multirow{3}{*}{Other} & D & 0.490 & 0.221 & 0.264 \\ \cline{2-5}
	 & F & 0.502 & 0.246 & 0.284 \\ \cline{2-5}
	 & DF & 0.517 & 0.260 & 0.276 \\ \hline
	\multirow{2}{*}{Asset combination} & D & 0.453 & 0.282 & 0.395 \\ \cline{2-5}
	 & DF & 0.565 & 0.200 & 0.274 \\ \hline
	Real estate & D & 0.506 & 0.217 & 0.308 \\ \hline
\end{tabular}
\label{tb:Ex-TFIDF}
\end{table}

\begin{table}[ht]
\centering
\caption{Detailed results in Ab-T5. D indicates domestic, F indicates foreign, and DF indicates domestic and foreign.}
\begin{tabular}{|c|c|r|r|r|} \hline
	\multicolumn{2}{|c}{} & \multicolumn{1}{|c}{ROUGE-1} & \multicolumn{1}{|c}{ROUGE-2} & \multicolumn{1}{|c|}{ROUGE-L} \\ \hline
	\multicolumn{2}{|c|}{All} & 0.704 & 0.548 & 0.595 \\ \hline
	\multirow{3}{*}{Stock} & D & 0.669 & 0.504 & 0.570 \\ \cline{2-5}
	 & F & 0.514 & 0.264 & 0.330 \\ \cline{2-5}
	 & DF & 0.470 & 0.177 & 0.253 \\ \hline
	\multirow{2}{*}{Bond} & F & 0.785 & 0.624 & 0.643 \\ \cline{2-5}
	 & DF & 0.596 & 0.446 & 0.499 \\ \hline
	\multirow{3}{*}{Other} & D & 0.730 & 0.592 & 0.633 \\ \cline{2-5}
	 & F & 0.728 & 0.579 & 0.624 \\ \cline{2-5}
	 & DF & 0.721 & 0.578 & 0.629 \\ \hline
	\multirow{2}{*}{Asset combination} & D & 0.405 & 0.209 & 0.316 \\ \cline{2-5}
	 & DF & 0.561 & 0.316 & 0.380 \\ \hline
	Real estate & D & 0.724 & 0.554 & 0.583 \\ \hline
\end{tabular}
\label{tb:Ab-T5}
\end{table}

\section{Discussion}
From Table \ref{tb:results}, abstractive summarization method Ab-T5 outperforms other extractive summarization methods.
Ab-T5 is superior for all of ROUGE-1, ROUGE-2, and ROUGE-L, indicating that the abstractive summarization is suitable for this task.
In particular, the ROUGE-2 and ROUGE-L values are very high compared to the extractive summarization results.
On the other hand, a comparison between extractive summarization results in a slightly higher Ex-TFIDF.
This result indicates that Ex-TFIDF was able to select more appropriate sentences as correct data.
In this research, we did not fine-tune BERT. 
Therefore, we believed that BERT was not able to calculate the appropriate vector to select the correct data.
If SBERT\cite{reimers2019sentence} were used instead of BERT, we consider that better results would be obtained.
Table \ref{tb:Ex-BERT}, Table \ref{tb:Ex-TFIDF}, and Table \ref{tb:Ab-T5} show that there are considerable differences in performance among fund types.
For example, for funds targeting foreign bonds, the value of the ROUGE-L is 0.643.
On the other hand, for funds that target domestic and foreign stocks, the value of ROUGE-L is 0.253.
These values difference is as large as 0.393.
From Table \ref{tb:Ab-T5}, we find that our method is good at bond and other, but not so good at stock and asset combination.
we believe that investment reports of bond and other are routine and suitable for automatic summarization.
On the other hand, we consider that investment reports for stock and other funds contain additional information that is not included in the monthly reports, making automatic summarization a bit difficult.
Therefore, we plan to construct a summarization model using additional information such as stock price and exchange rate.

\section{Related Works}
Related research on multi-document summarization includes the following papers.
Moro et al. proposed the probabilistic method based on the combination of three language models to tackle multi-document summarization in the medical domain\cite{moro2022discriminative}.
Liao et al. investigated the feasibility of utilizing Abstract Meaning Representation formalism for multi-document summarization\cite{liao-etal-2018-abstract}.
Fabbri et al. constructed Multi-News, the large-scale multi-document news summarization dataset\cite{fabbri-etal-2019-multi}.
Xiao et al. introduced PRIMERA, a pre-trained model for multi-document representation with a focus on summarization that reduces the need for dataset-specific architectures and large amounts of fine-tuning labeled data\cite{xiao2022primera}.
Nayeem et al. designed an abstractive fusion generation model at the sentence level, which jointly
performs sentence fusion and paraphrasing\cite{nayeem-etal-2018-abstractive}.
They applied their sentence-level model to implement an abstractive multi-document summarization system where documents usually contain a related set of sentences.
Liu et al. developed the neural summarization model, which can effectively process multiple input documents and distill abstractive summaries\cite{liu-lapata-2019-hierarchical}.
Li et al. develop a neural abstractive multi-document summarization model which can leverage explicit graph representations of documents to more effectively process multiple input documents and distill abstractive summaries\cite{li-etal-2020-leveraging-graph}.
Jin et al. proposed the multi-granularity interaction network to encode semantic representations for documents, sentences, and words\cite{jin-etal-2020-multi}.
Deyoung et al. released MSˆ2 (Multi-Document Summarization of Medical Studies), a dataset
of over 470k documents and 20K summaries derived from the scientiﬁc literature\cite{deyoung-etal-2021-ms}.     
As related work of extractive summarization, there is research by Cui et al.\cite{cui-hu-2021-sliding}.
They proposed extractive summarization that can summarize long-form documents without content loss.
Xu et al. proposed the neural network framework for extractive and compressive summarization \cite{xu-durrett-2019-neural}.
As related work of abstractive summarization, there is research by Nallapati et al.\cite{nallapati-etal-2016-abstractive}.
They applied the attentional encoder-decoder for the task of abstractive summarization with very promising results, outperforming state-of-the-art results signiﬁcantly on two different datasets.
Chen et al. proposed an accurate and fast summarization model that first selects
salient sentences and then rewrites them abstractively to generate a concise overall summary\cite{chen-bansal-2018-fast}.
Cohan et al. proposed the model for abstractive summarization of single, longer-form documents (e.g., research papers)\cite{cohan-etal-2018-discourse}.
Sharma et al. constructed BIGPATENT, the large-scale summarization dataset consisting of 1.3 million patent documents with human-written abstractive summaries\cite{sharma-etal-2019-bigpatent}.
Karn et al. proposed the extractive approach into a two-step RL-based summarization task (extractive-then-abstractive)\cite{karn-etal-2022-differentiable}.
Mao et al. proposed simple yet effective heuristics for oracle extraction as well as a consistency loss term, which encourages the extractor to approximate the averaged dynamic weights predicted by the generator\cite{mao-etal-2022-dyle}.

\section{Conclusion}
In this paper, we evaluated two extractive summarization methods and one abstractive summarization method on the investment report summarization from monthly reports.
The extractive summarization method is based on TFIDF or BERT model.
The abstractive summarization method is based on the T5 model.
From the evaluation results, we found that the abstractive summarization method outperformed our two extractive summarization methods. 
As feature work, we have a plan to construct a summarization model using additional information such as stock price and exchange rate.

\section*{Acknowledgments}
This work was supported by Daiwa Securities Group. 
This work was supported in part by JSPS KAKENHI Grant Number JP21K12010, the JST-Mirai Program Grant Number JPMJMI20B1, and JST-PRESTO Grant Number JPMJPR2267, Japan.

\bibliographystyle{IEEEtran}
\bibliography{list.bib}

\begin{table*}[ht]
  \caption{Examples of summarization results.}
  \begin{center}
  \small
  \begin{tabular}{p{17cm}}
  \hline \hline
  Correct Answer \\ \hline
  当作成期首より、ＦＯＭＣ（米国連邦公開市場委員会）において2019年内は利上げを行わない見通しであると示されたことなどを受けて、国債金利は低下傾向となりました。５月以降は、米中通商交渉への懸念が高まったことやＦＲＢ（米国連邦準備制度理事会）の利下げの実施などを背景に、国債金利の低下が続きました。米ドル建ての社債については、ＦＲＢの緩和的なスタンスなどを背景に、スプレッド（国債に対する上乗せ金利）は縮小圧力が強まる展開となりました。 \\
  (From the beginning of the current fiscal year, the FOMC (Federal Open Market Committee) indicated that it does not expect to raise interest rates in 2019, and interest rates on government bonds began to decline. Since May, government bond rates have continued to decline against the backdrop of heightened concerns over the U.S.-China trade negotiations and interest rate cuts by the Federal Reserve. As for U.S. dollar-denominated corporate bonds, spreads (interest rates on top of government bonds) came under increasing pressure to narrow against the backdrop of the Fed's accommodative stance and other factors.) \\
  \hline \hline
  Ab-T5 \\ \hline
  米国債券市場では、国債や社債の金利が低下(債券価格は上昇)しました。当作成期首より、C(米国連邦公開市場委員会)において2019年内は利上げを行わない見通しであると示されたことなどを受けて、国債金利は低下傾向となりました。5月以降は、米中通商交渉への懸念が高まったことや(米国連邦準備制度理事会)の利下げの実施などを背景に、国債金利の低下が続きました。米ドル建ての社債については、の緩和的なスタンスなどを背景に、スプレッド(国債に対する上乗せ金利)は縮小圧力が強まる展開となりました。\\
  (In the U.S. bond market, interest rates on government bonds and corporate bonds declined (bond prices rose). From the beginning of the period under review, government bond interest rates trended lower due to factors such as C (the U.S. Federal Open Market Committee's ) indication that it does not expect to raise interest rates within 2019, etc. From May, government bond interest rates continued to decline against the backdrop of heightened concerns over U.S.-China trade negotiations and the Federal Reserve Board's (US Fed) rate cut. Interest rates on U.S. government bonds continued to decline. In the U.S. dollar-denominated corporate bonds, spreads (interest rates on top of government bonds) came under increasing pressure to narrow due to the accommodative stance of the U.S. Federal Reserve Board (U.S. Federal Reserve Board).) \\
  \hline \hline
  Ex-TFIDF \\ \hline
  スプレッド（国債に対する上乗せ金利）は拡大米国債券市場では、米国による新たな対中関税の発表を契機として、米中貿易摩擦への懸念の高まりがリスク回避の動きにつながったことから、金利は低下しました。スプレッド（国債に対する上乗せ金利）は縮小米国債券市場では、米国の景況感指数や、中国における各種経済指標の改善を受けて、世界的な景気改善期待が高まったことから、国債金利は上昇しました。一方で、パウエルFRB（米国連邦準備制度理事会）議長の発言等を受けて利下げ期待が高まったことや、ECB（欧州中央銀行）の金融緩和期待が高まったことなどが金利低下要因となり、金利は前月末比でほぼ横ばいとなりました。スプレッド（国債に対する上乗せ金利）は拡大米国債券市場では、米国の中国に対する追加関税の引き上げを契機とした、米中対立の再燃を受けて、リスク回避的な動きの強まりから国債金利は低下しました。スプレッド（国債に対する上乗せ金利）は縮小米国債券市場では、米国の雇用統計等の経済指標の下振れに加えて、欧州の景況感が大幅に悪化し世界的な景気悪化観測を高めたことから、金利は低下しました。こうした環境下、米ドル建て社債については、投資家のリスク回避姿勢の強まりを背景にスプレッドは拡大したものの、金利が低下した銘柄も見られました。\\
  (In the U.S. bond market, interest rates declined as rising concerns about U.S.-China trade friction, triggered by the announcement of new U.S. tariffs on China, led to risk aversion. Spreads (interest rates on top of government bonds) narrowedIn the U.S. bond market, interest rates on government bonds rose as expectations of global economic improvement increased due to improvements in the U.S. business confidence index and various economic indicators in China. On the other hand, interest rates remained almost unchanged from the end of the previous month due to factors such as rising expectations of interest rate cuts following comments by Federal Reserve Chairman Jerome Powell and rising expectations of monetary easing by the European Central Bank (ECB), which led to a decline in interest rates. In the U.S. bond market, the U.S.-China conflict resurfaced, triggered by the U.S. raising additional tariffs on China, and government bond interest rates fell due to increased risk aversion. In the U.S. bond market, interest rates fell due to the downside of U.S. employment and other economic indicators, as well as a significant deterioration in business confidence in Europe, which raised expectations of a global economic downturn. In this environment, some issues of U.S. dollar-denominated corporate bonds saw interest rates decline, although spreads widened against the backdrop of investors' growing risk aversion.) \\
  \hline \hline
  \end{tabular}
  \end{center}
  \label{tb:outputs}
\end{table*}

\end{document}